%% file: RJwrapper.tex
\documentclass[a4paper]{report}
\usepackage[utf8]{inputenc}
\usepackage[T1]{fontenc}
\usepackage{RJournal}
\usepackage{amsmath,amssymb,array}
\usepackage{booktabs}

\providecommand{\tightlist}{%
  \setlength{\itemsep}{0pt}\setlength{\parskip}{0pt}}

\newlength{\cslhangindent}
\newlength{\csllabelwidth}
\newlength{\cslentryspacingunit} 
  {}%
  {\par}
\newenvironment{CSLReferences}[2] 
 {
  \setlength{\parindent}{0pt}
  \ifodd #1
  \let\oldpar\par
  \def\par{\hangindent=\cslhangindent\oldpar}
  \fi
  \setlength{\parskip}{#2\cslentryspacingunit}
 }%
 {}
\usepackage{calc}

\begin{document}

\sectionhead{Contributed research article}
\volume{XX}
\volnumber{YY}
\year{20ZZ}
\month{AAAA}

\begin{article}
  \input{li-dodwell-cook}
\end{article}

\end{document}

%% file: li-dodwell-cook.tex
\title{A Clustering Algorithm to Organize Satellite Hotspot Data for the Purpose of Tracking Bushfires Remotely}
\author{by Weihao Li, Emily Dodwell, and Dianne Cook}

\maketitle

\abstract{%
This paper proposes a spatiotemporal clustering algorithm and its implementation in the R package \CRANpkg{spotoroo}. This work is motivated by the catastrophic bushfires in Australia throughout the summer of 2019-2020 and made possible by the availability of satellite hotspot data. The algorithm is inspired by two existing spatiotemporal clustering algorithms but makes enhancements to cluster points spatially in conjunction with their movement across consecutive time periods. It also allows for the adjustment of key parameters, if required, for different locations and satellite data sources. Bushfire data from Victoria, Australia, is used to illustrate the algorithm and its use within the package.
}

\hypertarget{introduction}{%
\subsection{Introduction}\label{introduction}}

The 2019-2020 Australia bushfire season was catastrophic in the scale of damage caused to agricultural resources, property, infrastructure, and ecological systems. By the end of 2020, the devastation attributable to these Black Summer fires totalled 33 lives lost, almost 19 million hectares of land burned, over 3,000 homes destroyed and AUD \$1.7 billion in insurance losses, as well as an estimated 1 billion animals killed, including half of Kangaroo Island's population of koalas (Filkov et al. 2020). According to the Australian Government Bureau of Meteorology (2021), 2019 was the warmest year on record in Australia, and the period from 2013-2020 represents eight of the ten warmest years in recorded history. There is concern and expectation that impacts of climate change -- including more extreme temperatures, persistent drought, and changes in plant growth and landscape drying -- will worsen conditions for extreme bushfires (CSIRO and Australian Government Bureau of Meteorology 2020; Deb et al. 2020). Contributing to the problem is that dry lightning represents the main source of natural ignition, and fires that start in remote areas deep in the temperate forests are difficult to access and monitor (Abram et al. 2021). Therefore, opportunities to detect fire ignitions, monitor bushfire spread, and understand movement patterns in remote areas are important for developing effective strategies to mitigate bushfire impact.

Remote satellite data provides a potential solution to the challenge of active fire detection and monitoring. Development of algorithms that process satellite imagery into hotspots -- points that represent likely fires -- is an active area of research (see for example Giglio, Schroeder, and Justice (2016), Xu and Zhong (2017), Wickramasinghe et al. (2016), Jang et al. (2019)). Throughout this paper and the associated package, we make use of the Japan Aerospace Exploration Agency (JAXA) Himawari-8 satellite wildfire product (P-Tree System 2020) that identifies the location and fire radiative power (FRP) of hotspots across East Asia and Australia according to an algorithm developed by Kurihara et al. (2020). It contains records of 1,989,572 hotspots from October 2019 to March 2020 with a 0.02\(^o\) (\textasciitilde2km) spatial resolution and 10-minute temporal resolution. Detection of bushfire ignition and movement requires the clustering of satellite hotspots into meaningful clusters, which may then be considered in their entirety or summarized by a trajectory.

In this paper, we present a spatiotemporal clustering algorithm to organize hotspots into groups in order to estimate bushfire ignition locations and track bushfire movements over space and time. Inspired by two existing clustering algorithms, namely Density Based Spatial Clustering of Applications with Noise (DBSCAN) (Ester et al. 1996) and Fire Spread Reconstruction (FSR) (Loboda and Csiszar 2007), our algorithm adopts the notion of noise from DBSCAN, while drawing upon the fire movement dynamics presented in FSR. We generalize the latter's specification of spatiotemporal parameters, thereby providing an intuitive, straightforward, and extendable approach to the complex problem of bushfire identification and monitoring that may be applied to any satellite wildfire product. In clustering hotspots into bushfires of arbitrary shape and size, we aim to capture key fire behavior:

\begin{itemize}
\tightlist
\item
  fire evolution occurs only forwards in time,
\item
  fires can smolder undetectably for some time and then flare up again,
\item
  fires can merge with other bushfires, and
\item
  solitary hotspots, which we classify as noise, may not represent true fires or are otherwise very brief ignitions that do not spread.
\end{itemize}

\noindent This algorithm is implemented in the R package \CRANpkg{spotoroo}, available on CRAN. By enabling the user to cluster satellite hotspot data across space and time, this software provides the ability to relate findings to key factors in bushfire ignition (e.g.~weather, proximity to roads and campsites, and fuel sources) and patterns in their spread.

The core functionality of this spatiotemporal clustering algorithm determines whether a hotspot represents a new ignition point or a continuation of an existing bushfire by comparing and combining cluster membership information via incremental updates from one time window to the next. Our algorithm first slices the hotspot data by its temporal dimension into fixed time intervals, according to a user-defined time step. Doing so divides the overall spatiotemporal clustering task into many smaller spatial clustering tasks that may be completed in parallel. Within each time window, which can be considered a static snapshot of hotspots observed across a user-specified number of time intervals, hotspots that fall within the threshold of a user-defined spatial metric of each other are joined into a cluster. Then, proceeding sequentially, we identify whether or not a hotspot was observed in the previous time window. If so, it retains its cluster membership from the previous time window; if not, the hotspot adopts the membership of the nearest hotspot with which it has been clustered. If no such neighbor exists, a hotspot represents the start of a new fire. It is important to note that each hotspot does not necessarily represent an individual fire, so those clusters that do not pass the threshold of a minimum number of hotspots or exist for a minimum amount of time are labelled noise.

As emphasized by Kisilevich et al. (2009), the selection of spatial resolution and time granularity -- and relevance of domain knowledge in their choice -- are imperative to the understanding and interpretation of resulting clusters. These choices can influence the shape and number of clusters discovered. In the case of satellite hotspot data, these parameters depend on the spatial resolution and temporal frequency at which images are captured. We suggest that parameter tuning can be assessed using a visual heuristic, which also enables selection of appropriate values in a unit-free manner, independent of a satellite's spatial and temporal resolutions.

This paper is organized as follows. The next section provides an introduction to the literature on spatiotemporal clustering and applications to bushfire modeling. Section \protect\hyperlink{algorithm}{Algorithm} details the steps of the clustering algorithm, and Section \protect\hyperlink{package}{Package} introduces its implementation in \CRANpkg{spotoroo} on CRAN, including demonstration of the package's key functions and visualization capabilities. We illustrate the clustering algorithm's functionality to study bushfire ignition and spread in Victoria, Australia throughout the 2020 bushfire season in \protect\hyperlink{application}{Application}, and describe a visual heuristic to inform parameter selection. Finally, we give a brief conclusion of the paper and discuss potential opportunities for use of the clustering algorithm.

\hypertarget{background}{%
\subsection{Background}\label{background}}

Han, Kamber, and Pei (2012) overviews clustering methods and groups algorithms into five types: partitioning, hierarchical, density-based, grid-based, and model-based methods. Clustering of hotspot data lends itself nicely to hierarchical and density-based methods because they allow for the identification of clusters of various shapes and sizes without requiring that the user pre-specify the number of clusters. Particularly, density-based methods have the notion of noise, which is convenient for eliminating non-fire events from the clustering result. We therefore focus on a review of density-based methods and refer the reader to Han, Kamber, and Pei (2012) for algorithms in other categories and Kisilevich et al. (2009) for appropriate extensions to spatiotemporal data.

\hypertarget{spatiotemporal-clustering-based-on-dbscan}{%
\subsubsection{Spatiotemporal clustering based on DBSCAN}\label{spatiotemporal-clustering-based-on-dbscan}}

Density-based methods separate regions constituting a high density of points from low-density regions by identifying pairwise distances between points, and then requiring that a threshold for their grouping be satisfied (Han, Kamber, and Pei 2012). Density Based Spatial Clustering of Applications with Noise (DBSCAN) (Ester et al. 1996) is an implementation of this methodology designed to address three challenges of clustering algorithms: (1) requirements of domain knowledge to determine the hyperparameters, (2) arbitrary shape of clusters and (3) computational efficiency. DBSCAN defines a maximum radius to construct a neighborhood around each point. It distinguishes between a core point, for which the number of points that fall in its neighborhood meets a minimum threshold, and a boundary point, whose neighborhood does not meet this threshold, but can be reached via overlapping neighborhoods extending from the core point. Intersecting neighborhoods are defined to be a cluster, while points that cannot be assigned to a cluster are identified as noise. DBSCAN also provides a heuristic to inform selection of threshold and cluster size.

What is often identified as a limitation of DBSCAN -- its inability to differentiate between clusters of different densities and those adjacent to each other (Birant and Kut 2007) -- is of less concern for the application to satellite data, which by nature is a set of points corresponding to the equidistant center of pixels on grid of latitudes and longitudes. However, its application to spatiotemporal clustering problems, which contain at least three components -- spatial location (e.g.~latitude and longitude) and time -- requires specification of temporal granularity and treatment of temporal similarity (Kisilevich et al. 2009). As such, several extensions to DBSCAN's spatial clustering functionality have been proposed for spatiotemporal clustering solutions.

ST-DBSCAN (Birant and Kut 2007) was developed as an extension of DBSCAN's functionality to cluster points according to their non-spatial, spatial, and temporal attributes, and it simultaneously addresses DBSCAN's limitations regarding identification of clusters of varying densities and differentiation of adjacent clusters. ST-DBSCAN also introduces a second metric that considers similarity of variables associated with temporal neighbors.

Another spatiotemporal extension of DBSCAN called MC1 was developed by Kalnis, Mamoulis, and Bakiras (2005). The authors slice the data into snapshots along the time dimension and define moving clusters using the percentage of common objects shared by clusters in two consecutive snapshots.
The clustering result is obtained by linking the outputs of DBSCAN applied on each snapshot with the definition of moving clusters. This algorithm enables tracking of moving clusters such as a convoy of cars moving in a city efficiently. However, the definition of moving clusters assumes that data points are individual objects with unique IDs that can move throughout time, which does not translate to hotspots.

\hypertarget{spatiotemporal-clustering-with-fsr}{%
\subsubsection{Spatiotemporal clustering with FSR}\label{spatiotemporal-clustering-with-fsr}}

Satellite hotspot data has been effectively clustered and visualized using DBSCAN, as demonstrated in studies by Nisa, Andrianto, and Mardhiyyah (2014) and Hermawati and Sitanggang (2016). However, it should be noted that DBSCAN does not enable the tracking of fire ignition and movement over time.

FSR (Loboda and Csiszar 2007) addresses this limitation with a focus on fire dynamics, which aim to characterize the ignition location, spatial progression, and rate of movement of individual fire events. It introduces what is effectively a hierarchical-based clustering algorithm to identify fire spread in the Russian boreal forest based on active fire detection from MODIS (Moderate Resolution Imaging Spectroradiometer), which has a temporal frequency of six hours. The algorithm proposed by the authors constructs a tree based on three rules: (1) the earliest observed hotspot is the root of the tree, (2) any node is within a 2.5km radius from its parent and (3) any node is observed no later than four days from its parent. When the tree is closed and there are still unassigned hotspots, the algorithm continues at the earliest unassigned hotspot to construct a new tree. Finally, each tree is a cluster, and the earliest observed hotspot(s) is defined as the ignition point (i.e.~a cluster may have multiple ignition points).

The selection of parameters for FSR is tailored to the specific region and data product being utilized. As a result, these parameter settings cannot be immediately generalized or applied to other sources of satellite hotspot data.

\hypertarget{limitations-of-existing-methods-for-the-purposes-of-bushfire-monitoring}{%
\subsubsection{Limitations of existing methods for the purposes of bushfire monitoring}\label{limitations-of-existing-methods-for-the-purposes-of-bushfire-monitoring}}

When considered in the context of clustering hotspot data, the existing methods discussed in the previous section are not ideal. Here we clarify these limitations that have led to different choices in our algorithm, particularly in combining the clustering results of two consecutive time periods.

ST-DBSCAN's consideration of the similarity of variables associated with consecutive time periods adds an unnecessary complexity for the spatiotemporal clustering of hotspots, for which we focus on three variables (latitude, longitude, time). It also does not guarantee that clusters are processed temporally, which is required to determine bushfire ignition sites. For this reason, we have developed a different handling of the temporal variable for our hotspot clustering.

MC1 applies DBSCAN to each snapshot of data, which has the potential to treat some fire ignitions as noise. It is possible for a fire ignition to be initially identified by only a few hotspots, which may not be enough to meet DBSCAN's density constraint. While we believe it is important to introduce the concept of noise to filter out non-fire events, their identification and removal should not occur at each temporal snapshot for this reason. We have therefore implemented the noise filter differently.

We also consider how best to handle the spatial convergence of clusters, that is, when two or more clusters from a previous time period merge together in the current time period. In such situations, both MC1 and FSR assign only one membership to the cluster of hotspots in the current time period. The consequence of such treatment is twofold: spatial coverage of one fire may increase dramatically in a short amount of time, which may not accurately reflect the natural speed of a bushfire's spread, convergence or shifts in directions. This is not necessarily appropriate in light of our goal of identifying bushfire ignition locations. For this reason, and in contrast to MC1 and FSR, our algorithm inherently enables tracking of individual fires through their potential merging by keeping separate cluster ids in the current time period. This is because cluster membership is informed by a hotspot's appearance in previous time periods. In the event that multiple hotspots are observed in a cluster's first appearance, we calculate their centroid and record it as the ignition location.

\hypertarget{algorithm}{%
\subsection{Algorithm}\label{algorithm}}

Our spatiotemporal clustering algorithm consists of four steps: (1) divide hotspots into equal time intervals, (2) cluster hotspots spatially across a series of time intervals, which we will refer to as a time window, (3) combine cluster information between consecutive time windows, and (4) identify hotspots that represent noise so they can be filtered out. These four steps are described in detail in this section.

\hypertarget{divide-hotspots-into-intervals}{%
\subsubsection{1. Divide hotspots into intervals}\label{divide-hotspots-into-intervals}}

Because fires progress through time and our aim is to determine the ignition point of any particular bushfire, time is a critical component for determining a fire's starting point and tracking its subsequent progression. To manage this analysis, it is convenient to partition time into intervals to support the spatial clustering of hotspots and eventual combination of results with those of future intervals.

Selection of an appropriate time interval for clustering hotspot data represents a balance between maintaining the resolution of the original satellite data (which may record observations every five minutes or every hour or every four hours, for example) and computational efficiency of the algorithm. For example, a time interval of half an hour produces twice the number of iterations of the spatial clustering algorithm than a time interval of an hour; computation time is approximately linear with number of iterations. Himawari-8 hotspot data is recorded every 10 minutes; using this original temporal granularity has the potential to introduce significant noise, and would represent many more iterations of the clustering algorithm. For mitigation of this noise, we may choose an interval of 60 minutes = 1 hour. We then maintain an overall index of each hour across our entire data set, denoted as \(t=1,...,T\) where T is the total number of hours in the data set.

A hotspot is a likely indication of a burning fire, although it is possible that a hotspot may disappear between two consecutive intervals, perhaps due to cloud cover or moisture that dampens the fire such that it becomes temporarily undetectable. The algorithm must take these considerations into account when combining information across consecutive time intervals. We need a parameter to measure how long a fire may burn undetected even if we don't see a hotspot; that is, the amount of time a bushfire is considered to be active but unaccounted for in the hotspot data.

For this reason, the parameter \(activeTime\) is defined to reflect in intervals the maximum amount of time a fire may stay smoldering but undetectable by satellite before flaring up again. For example, if it is reasonable to assume that a bushfire not observed for 24 hours has burned out, the implication on \(activeTime\) is as follows: if time is divided into half hour intervals, the parameter \(activeTime\) will be 48, whereas with hour intervals, \(activeTime\) is 24. The \(activeTime\) parameter is a unitless integer.

Let \(\boldsymbol{S}_t\) be a series of time intervals defined by

\[\boldsymbol{S}_t = [max(1,t-activeTime),t]\quad t = 1,2,...,T,\]

\noindent where \(max(.)\) is the maximum function, \(t\) is an integer time index, and \(T\) is the integer length of the time window.

Suppose the data set contains \(T\) hours of hotspot data, our time interval is one hour, and \(activeTime = 24\). This produces a sequence of time windows
\(\boldsymbol{S}_1,\boldsymbol{S}_2,\ldots,\boldsymbol{S}_{T}\), where

\begin{align*}
\boldsymbol{S}_1 &= [1,1],\\
\boldsymbol{S}_2 &= [1,2],\\
&...\\
\boldsymbol{S}_{25} &= [1,25],\\
\boldsymbol{S}_{26} &= [2,26],\\
&...\\
\boldsymbol{S}_{47} &= [23,47],\\
\boldsymbol{S}_{48} &= [24,48],\\
&...\\
\boldsymbol{S}_{T} &= [T-24,T].
\end{align*}

\hypertarget{cluster-hotspots-spatially-within-each-time-window}{%
\subsubsection{2. Cluster hotspots spatially within each time window}\label{cluster-hotspots-spatially-within-each-time-window}}

We next cluster the hotspots within each time window. Each time window represents a static unified view of all hotspots observed across 25 consecutive time intervals (as of \(\boldsymbol{S}_{25}\)). The clustering within a time window disregards the specific hourly time index associated with each observed hotspot to focus exclusively on the spatial relationship between them. A hierarchical-based method with a maximum distance stopping criterion \(adjDist\) is applied.

The parameter \(adjDist\) is used to represent the maximum distance between two adjacent hotspots, which are connected to form a graph. For example, the Himawari-8 hotspot data collection has 0.02\(^o\) resolution, which corresponds to approximately 2000m. Thus, \(adjDist\) is set to 3000m to indicate diagonal distance between observations. Every connected component of the graph is then considered as an individual cluster.

Given \(adjDist>0~m\) and a window \(\boldsymbol{S}_t\), the algorithm performs the following substeps:

\begin{enumerate}
\def\labelenumi{(\alph{enumi})}
\item
  Append a randomly selected hotspot \(h_i\) to a empty list \(\boldsymbol{L}\), where \(h_i\) is the \(i\)th hotspot in the time window \(\boldsymbol{S}_t\). Let point \(\boldsymbol{P}\) be the first item of the list \(\boldsymbol{L}\).
\item
  For every \(h_i \notin \boldsymbol{L}\), if \(geodesic(h_i, \boldsymbol{P})\leq adjDist\), append \(h_i\) to the list \(\boldsymbol{L}\).
\item
  Let point \(\boldsymbol{P}\) be the next item of the list \(\boldsymbol{L}\).
\item
  Repeat (b) and (c) until the point \(\boldsymbol{P}\) reaches the end of the list \(\boldsymbol{L}\).
\item
  For all hotspots \(h_i \in \boldsymbol{L}\), assign a new membership to them. Remove these hotspots from the time window \(\boldsymbol{S}_t\). Repeat (a) to (e) until time window \(\boldsymbol{S}_t\) is empty.
\item
  Recover the time window \(\boldsymbol{S}_t\) and record the memberships.
\end{enumerate}

\begin{figure}

{\centering \includegraphics[width=1\linewidth]{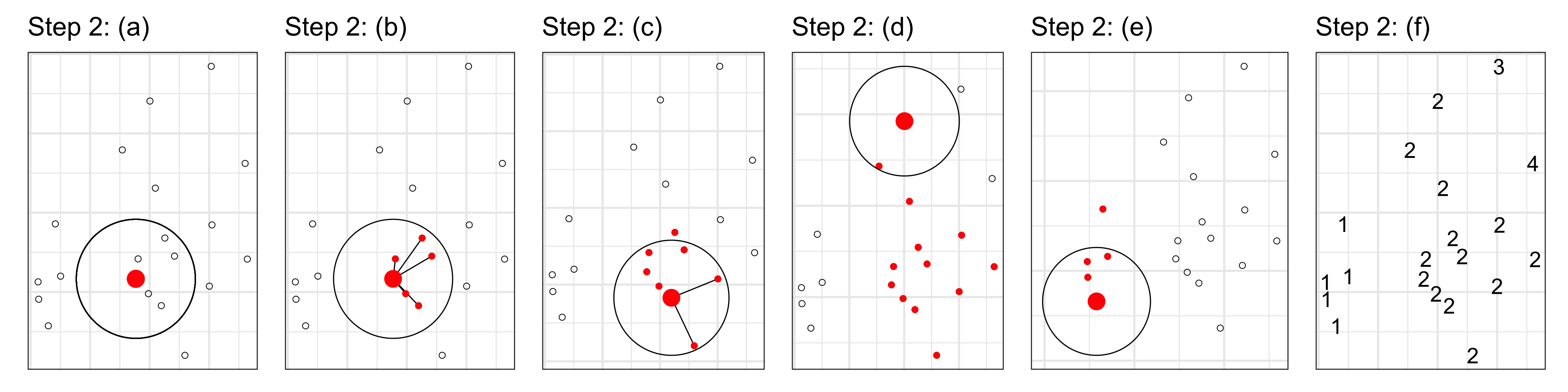} 

}

\caption{Illustration showing Step 2 of the clustering algorithm on a sample of 20 hotspots in one time window $\boldsymbol{S}_t$. Initially (a), a hotspot is selected randomly ($\boldsymbol{P}$) in order to seed a cluster. The circle indicates the maximum neighborhood distance ($adjDist$). Nearby hotspots as shown in red are clustered with $\boldsymbol{P}$ (b) to initialize list $\boldsymbol{L}$. The neighborhood is moved following every point in that collected list $\boldsymbol{L}$ and new observations are added (c), until there no more points that can be grouped (d). Then a new hotspot is selected external to the existing cluster, and the process is repeated (e). At the end, all the hotspots will be clustered (f).}\label{fig:step2figs}
\end{figure}

\noindent Figure \ref{fig:step2figs} provides an example of this step.

\hypertarget{update-memberships-for-next-time-window}{%
\subsubsection{3. Update memberships for next time window}\label{update-memberships-for-next-time-window}}

\begin{figure}

{\centering \includegraphics[width=1\linewidth]{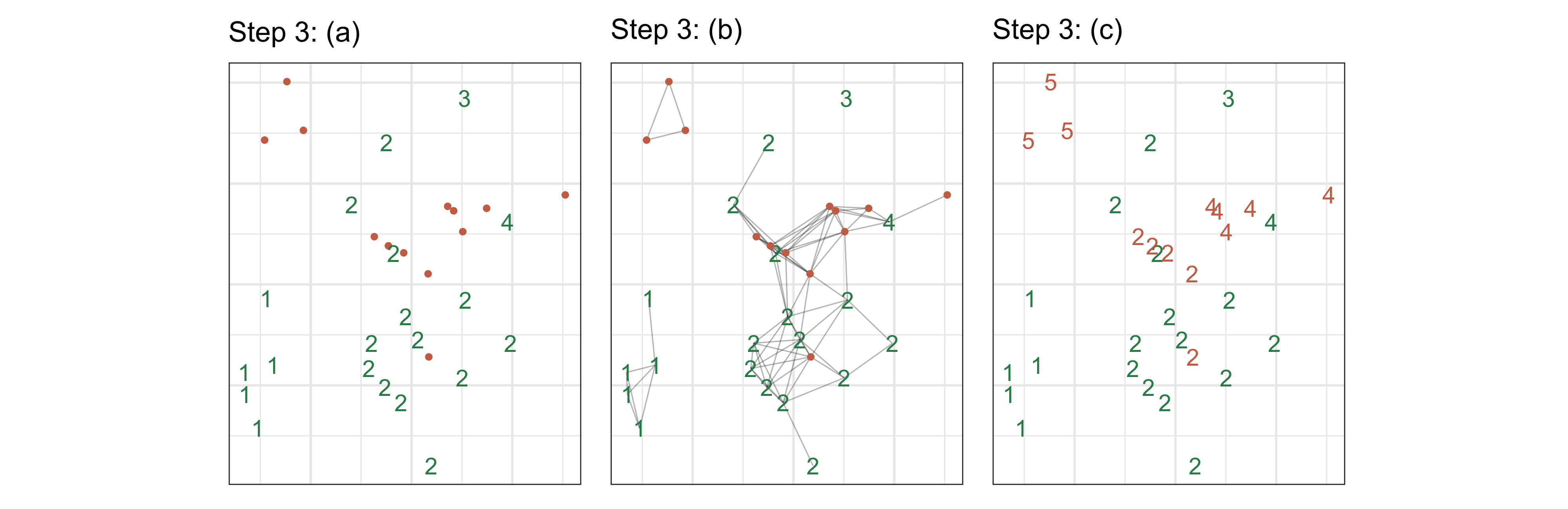} 

}

\caption{Illustration of clustering Step 3, which involves combining results from one time window to the next. There are 33 hotspots at $\boldsymbol{S}_t$, where 20 (green) of them have been previously clustered at $\boldsymbol{S}_{t-1}$ (Figure 1 f) and 13 (orange) of them are new hotspots. The connected graph show the clustering in this time window. Hotspots previously clustered at $\boldsymbol{S}_{t-1}$ keep their cluster labels. The 13 new hotspots are assigned labels of the nearest hotspot's cluster label. This might mean that a big cluster $\boldsymbol{S}_t$ (indicated by the graph) would be split back into two, if it corresponded to two clusters at $\boldsymbol{S}_{t-1}$ (e.g. clusters 2, 4). New clusters of hotspots are assigned a new label (e.g. cluster 5).}\label{fig:step3figs}
\end{figure}

Step 3 assigns membership to hotspots in the next time window (\(t=2, ..., T\) ) based on memberships in the previous window, as illustrated by Figure \ref{fig:step3figs}. The algorithm performs the following substeps:

\begin{enumerate}
\def\labelenumi{(\alph{enumi})}
\item
  Collect the hotspots, \(h_i\), belonging to \(\boldsymbol{S}_{t}\), where some may have also been present in \(\boldsymbol{S}_{t-1}\), into the set \(\boldsymbol{H}_t = \{h_1,h_2,...\}\). Hotspots that were present in \(\boldsymbol{S}_{t-1}\) are labelled, and new hotspots are unlabeled.
\item
  Construct the connectivity graph, based on \(adjDist\), measuring proximity between all hotspots in \(\boldsymbol{H}_t\). There may be some hotspots from multiple \(\boldsymbol{S}_{t-1}\) clusters connected together here, due to new hotspots appearing between previous hotspots. This is resolved at the next step.
\item
  For all unlabeled hotspots, use the label of the closest labelled point, if it is connected in the graph. Any unlabeled hotspots not connected to labelled hotspots are assigned a new label (cluster 5 in Figure \ref{fig:step3figs}). Also, any connected graph where memberships from multiple \(\boldsymbol{S}_{t-1}\) clusters occur are effectively split into the same multiple clusters (clusters 2 and 4 in Figure \ref{fig:step3figs}). This would correspond to two existing fires burning into each other, but they keep their original label. These fires might die at this time, or they might progress in different directions.
\end{enumerate}

This consideration of cluster memberships in consecutive time windows enables us to capture the long-term behavior of individual fires, namely whether they go undetected for 24 hours or more. Intuitively, a fire is undetected for 24 hours if a cluster is represented in \(\boldsymbol{S}_{t-1}\) but not \(\boldsymbol{S}_{t}\); that is, all hotspots in the cluster as of time window \(\boldsymbol{S}_{t-1}\) were from time index \(t-25\). By moving through the temporal sequence of time windows this way, we gradually identify the ignition time and location of new fires by considering hotspot activity in a region over the last 24 hours.

\hypertarget{handle-noise-in-the-clustering-result}{%
\subsubsection{4. Handle noise in the clustering result}\label{handle-noise-in-the-clustering-result}}

After performing step 3, all hotspots will have been assigned a membership label. However, there may exist small clusters, where small refers to number of associated hotspots and/or length of time for which a cluster is observed. These small clusters are less important for bushfire monitoring due to their limited spread, and we may be less certain that they represent true fires. To address this issue, we provide a noise filter as the last step.

Parameter \(minPts\) specifies the minimum number of hotspots a cluster can contain and parameter \(minTime\) specifies the minimum amount of time for which a cluster can exist and still be considered a bushfire. Any cluster that does not satisfy these two conditions will be reassigned membership label \(-1\) to indicate they represent noise.

\hypertarget{result}{%
\subsubsection{Result}\label{result}}

The result of the spatiotemporal clustering algorithm applied to the hotspot data is a vector of memberships with length equal to the number of observations in the data.

\hypertarget{package}{%
\subsection{Package}\label{package}}

Our spatiotemporal clustering algorithm detailed in the previous section is implemented in the R package \CRANpkg{spotoroo}, which is available for download on CRAN:

\begin{verbatim}
install.packages("spotoroo")
\end{verbatim}

The development version can be installed from Github:

\begin{verbatim}
devtools::install_github("TengMCing/spotoroo")
\end{verbatim}

The following demonstration of the package's functionality assumes \CRANpkg{spotoroo} and \CRANpkg{dplyr} have been loaded. We also want to load the built-in data set \texttt{hotspots}.

\begin{verbatim}
library(spotoroo)
library(dplyr)
data(hotspots)
\end{verbatim}

The \texttt{hotspots} data is a subset of the original Himawari-8 wildfire product and contains 1070 observations observed in the state of Victoria in Australia from December 2019 through February 2020.

To create this illustrative data set, records from the original October 2019-March 2020 data set were filtered to include only those hotspots in the boundary of Victoria's borders with an irradiance over 100 watts per square meter. This threshold of the hotspot's intensity is proposed by landscape ecologist and spatial scientist Dr.~Grant Williamson (2020) to reduce the likelihood of including hotspots that do not represent true fires in our analysis. These two processing steps capture 75,936 observations, for which we preserve fields longitude, latitude, and observed date and time; we sample approximately 1\% of these records for the package's final \texttt{hotspots} data set.

\hypertarget{performing-spatiotemporal-cluster-analysis}{%
\subsubsection{Performing spatiotemporal cluster analysis}\label{performing-spatiotemporal-cluster-analysis}}

The main function of this package is \texttt{hotspot\_cluster()}, which implements the spatiotemporal clustering algorithm on satellite hotspot data input by the user.

This function accepts a data set of hotspots (\texttt{hotspots}) which must contain fields corresponding to longitude (\texttt{lon}), latitude (\texttt{lat}), and observed time (\texttt{obsTime}). Arguments \texttt{activeTime}, \texttt{adjDist}, \texttt{minPts}, and \texttt{minTime} were previously defined in the \protect\hyperlink{algorithm}{Algorithm} section, and represent parameters for the algorithm's functionality. Arguments \texttt{timeUnit} and \texttt{timestep} represent the conversion from a hotspot's observed time to its corresponding integer time index.

We illustrate the use of function \texttt{hotspot\_cluster()} below. Our \texttt{hotspots} data set has columns with names that correspond to the first four arguments. Parameters \texttt{timeUnit} and \texttt{timeStep} define the difference between two adjacent time indices as \(1\) hour, and \texttt{activeTime} is \(24\) time indexes, \texttt{adjDist} is \(3000\) meters, \texttt{minPts} is \(4\) hotspots, and \texttt{minTime} is \(3\) time indices. That is, we consider any cluster that lasts longer than \(3\) hours and contains at least \(4\) hotspots to be a bushfire, although since the definition of a bushfire may vary according to geography, these parameters may be adjusted accordingly by the user.

\begin{verbatim}
result <- hotspot_cluster(hotspots = hotspots,
                          lon = "lon",
                          lat = "lat",
                          obsTime = "obsTime",
                          activeTime = 24,
                          adjDist = 3000,
                          minPts = 4,
                          minTime = 3,
                          timeUnit = "h",
                          timeStep = 1)
\end{verbatim}

This function returns a \texttt{spotoroo} object, which is a three-object \texttt{list} containing two \texttt{data.frame}s -- \texttt{hotspots} and \texttt{ignition} -- and a \texttt{list} called \texttt{setting}. Printing this output tells us that of the 1070 original hotspots, 10 were identified as noise points, and the remaining 1060 were clustered into 6 clusters.

\begin{verbatim}
result
\end{verbatim}

\begin{verbatim}
#> i spotoroo object: 6 clusters | 1070 hot spots (including noise points)
\end{verbatim}

Within \texttt{result}, the first item is a \texttt{hotspots} data frame. This contains the three fields from our original input data (namely, \texttt{lon}, \texttt{lat}, and \texttt{obsTime}), as well as each hotspot's assigned time index (\texttt{timeID}) and cluster \texttt{membership}. Each hotspot is also associated with the calculated distance from and time since its associated fire's ignition, \texttt{distToIgnition} and \texttt{timeFromIgnition}, respectively. Any hotspot with a membership of \(-1\) is identified as noise (\texttt{noise} == \texttt{TRUE}).

Our original hotspot data was recorded at 10-minute intervals, and we defined each time interval to be one hour. A hotspot's \texttt{timeID} is assigned starting from the first observed timestamp in the data. For example, \texttt{2019-12-29\ 13:10:00} is the earliest \texttt{obsTime} in \texttt{hotspots} and begins the hour for which \texttt{timeID\ ==\ 1}; any hotspot observed from \texttt{2019-12-29\ 14:10:00} to \texttt{2019-12-29\ 15:00} is assigned \texttt{timeID\ ==\ 2}, and so on. This enumeration continues through the last hour for which we observe hotspots in the data.

\begin{verbatim}
result$hotspots %>% arrange(obsTime) %>% glimpse()
\end{verbatim}

\begin{verbatim}
#> Rows: 1,070
#> Columns: 10
#> $ lon                  <dbl> 149.30, 149.30, 149.32, 149.30, 149.30, 149.32, 1~
#> $ lat                  <dbl> -37.75999, -37.78000, -37.78000, -37.75999, -37.7~
#> $ obsTime              <dttm> 2019-12-29 13:10:00, 2019-12-29 13:10:00, 2019-1~
#> $ timeID               <int> 1, 1, 1, 2, 2, 2, 2, 2, 2, 3, 4, 4, 5, 5, 7, 10, ~
#> $ membership           <dbl> 1, 1, 1, 1, 1, 1, 1, 1, 1, 1, 1, 1, 1, 1, 1, 1, 1~
#> $ noise                <lgl> FALSE, FALSE, FALSE, FALSE, FALSE, FALSE, FALSE, ~
#> $ distToIgnition       <dbl> 1111.885, 1111.885, 2080.914, 1111.885, 1111.885,~
#> $ distToIgnitionUnit   <chr> "m", "m", "m", "m", "m", "m", "m", "m", "m", "m",~
#> $ timeFromIgnition     <drtn> 0.0000000 hours, 0.0000000 hours, 0.3333333 hour~
#> $ timeFromIgnitionUnit <chr> "h", "h", "h", "h", "h", "h", "h", "h", "h", "h",~
\end{verbatim}

The second item within \texttt{result} is the \texttt{ignition} data set, which contains one row for each cluster identified by the algorithm. Each cluster's row captures information regarding its ignition location, observed time of ignition, number of hotspots in the cluster, and for how long the cluster was observed. The coordinates of each ignition point represent the earliest observed hotspot associated with the cluster, or the calculated centroid if there are multiple hotspots observed in the first time index associated with the cluster.

\begin{verbatim}
glimpse(result$ignition)
\end{verbatim}

\begin{verbatim}
#> Rows: 6
#> Columns: 8
#> $ membership         <int> 1, 2, 3, 4, 5, 6
#> $ lon                <dbl> 149.3000, 146.7200, 149.0200, 149.1600, 146.7067, 1~
#> $ lat                <dbl> -37.77000, -36.84000, -37.42000, -37.29000, -36.993~
#> $ obsTime            <dttm> 2019-12-29 13:10:00, 2020-01-08 01:40:00, 2020-01-0~
#> $ timeID             <int> 1, 229, 258, 280, 327, 859
#> $ obsInCluster       <dbl> 146, 165, 126, 256, 111, 256
#> $ clusterTimeLen     <drtn> 116.1667 hours, 148.3333 hours, 146.3333 hours, 12~
#> $ clusterTimeLenUnit <chr> "h", "h", "h", "h", "h", "h"
\end{verbatim}

The final object contained in \texttt{result} is a list that records the user-input values for the algorithm's parameters.

\begin{verbatim}
result$setting
\end{verbatim}

\begin{verbatim}
#> $activeTime
#> [1] 24
#> 
#> $adjDist
#> [1] 3000
#> 
#> $minPts
#> [1] 4
#> 
#> $ignitionCenter
#> [1] "mean"
#> 
#> $timeUnit
#> [1] "h"
#> 
#> $timeStep
#> [1] 1
\end{verbatim}

The user can run the function \texttt{summary()} to return key distributions that summarize the clustering results. For clusters, it captures the distribution of the number of hotspots in a cluster and the duration of the cluster (in hours). Similarly, for hotspots, it captures the distribution of their distance to ignition, both spatially (in meters) and temporally (in hours). It also notes the number of noise observations identified by the clustering algorithm.

\hypertarget{extracting-a-subset-of-clusters}{%
\subsubsection{Extracting a subset of clusters}\label{extracting-a-subset-of-clusters}}

Some users may prefer to engage with a data frame of the results, rather than parsing the \texttt{spotoroo} list object described above. For this purpose, the package provides a function \texttt{extract\_fire()}, which converts a \texttt{spotoroo} object to a \texttt{data.frame} by collapsing the \texttt{hotspots} and \texttt{ignition} objects. A new field, \texttt{type}, records whether a row in this data frame represents a hotspot, ignition point, or noise point. Cluster information is available for each hotspot that was not identified to be noise. Users can include or disregard noise points by toggling the argument \texttt{noise\ =\ TRUE}.

\begin{verbatim}
all_fires <- extract_fire(result, noise = TRUE)
all_fires %>% arrange(obsTime) %>% glimpse()
\end{verbatim}

\begin{verbatim}
#> Rows: 1,076
#> Columns: 14
#> $ lon                  <dbl> 149.30, 149.30, 149.30, 149.32, 149.30, 149.30, 1~
#> $ lat                  <dbl> -37.75999, -37.78000, -37.77000, -37.78000, -37.7~
#> $ obsTime              <dttm> 2019-12-29 13:10:00, 2019-12-29 13:10:00, 2019-1~
#> $ timeID               <int> 1, 1, 1, 1, 2, 2, 2, 2, 2, 2, 3, 4, 4, 5, 5, 7, 1~
#> $ membership           <dbl> 1, 1, 1, 1, 1, 1, 1, 1, 1, 1, 1, 1, 1, 1, 1, 1, 1~
#> $ noise                <lgl> FALSE, FALSE, FALSE, FALSE, FALSE, FALSE, FALSE, ~
#> $ distToIgnition       <dbl> 1111.885, 1111.885, 0.000, 2080.914, 1111.885, 11~
#> $ distToIgnitionUnit   <chr> "m", "m", "m", "m", "m", "m", "m", "m", "m", "m",~
#> $ timeFromIgnition     <drtn> 0.0000000 hours, 0.0000000 hours, 0.0000000 hour~
#> $ timeFromIgnitionUnit <chr> "h", "h", "h", "h", "h", "h", "h", "h", "h", "h",~
#> $ type                 <chr> "hotspot", "hotspot", "ignition", "hotspot", "hot~
#> $ obsInCluster         <dbl> 146, 146, 146, 146, 146, 146, 146, 146, 146, 146,~
#> $ clusterTimeLen       <drtn> 116.1667 hours, 116.1667 hours, 116.1667 hours, ~
#> $ clusterTimeLenUnit   <chr> "h", "h", "h", "h", "h", "h", "h", "h", "h", "h",~
\end{verbatim}

By providing a vector of indices to the argument \texttt{cluster}, the function will extract the corresponding clusters from the clustering result.

\begin{verbatim}
fire_1_and_2 <- extract_fire(result, cluster = c(1, 2), noise = FALSE)
\end{verbatim}

\hypertarget{visualizing-the-clustering-result}{%
\subsubsection{Visualizing the clustering result}\label{visualizing-the-clustering-result}}

The package provides three basic methods to visualize the clustering results, all of which can be produced by the function \texttt{plot()}. Users can draw advanced graphics using the results provided by the algorithm.

\hypertarget{default-plot-for-visualizing-the-spatial-distribution-of-clusters}{%
\paragraph{Default plot for visualizing the spatial distribution of clusters}\label{default-plot-for-visualizing-the-spatial-distribution-of-clusters}}

\begin{figure}

{\centering \includegraphics[width=1\linewidth]{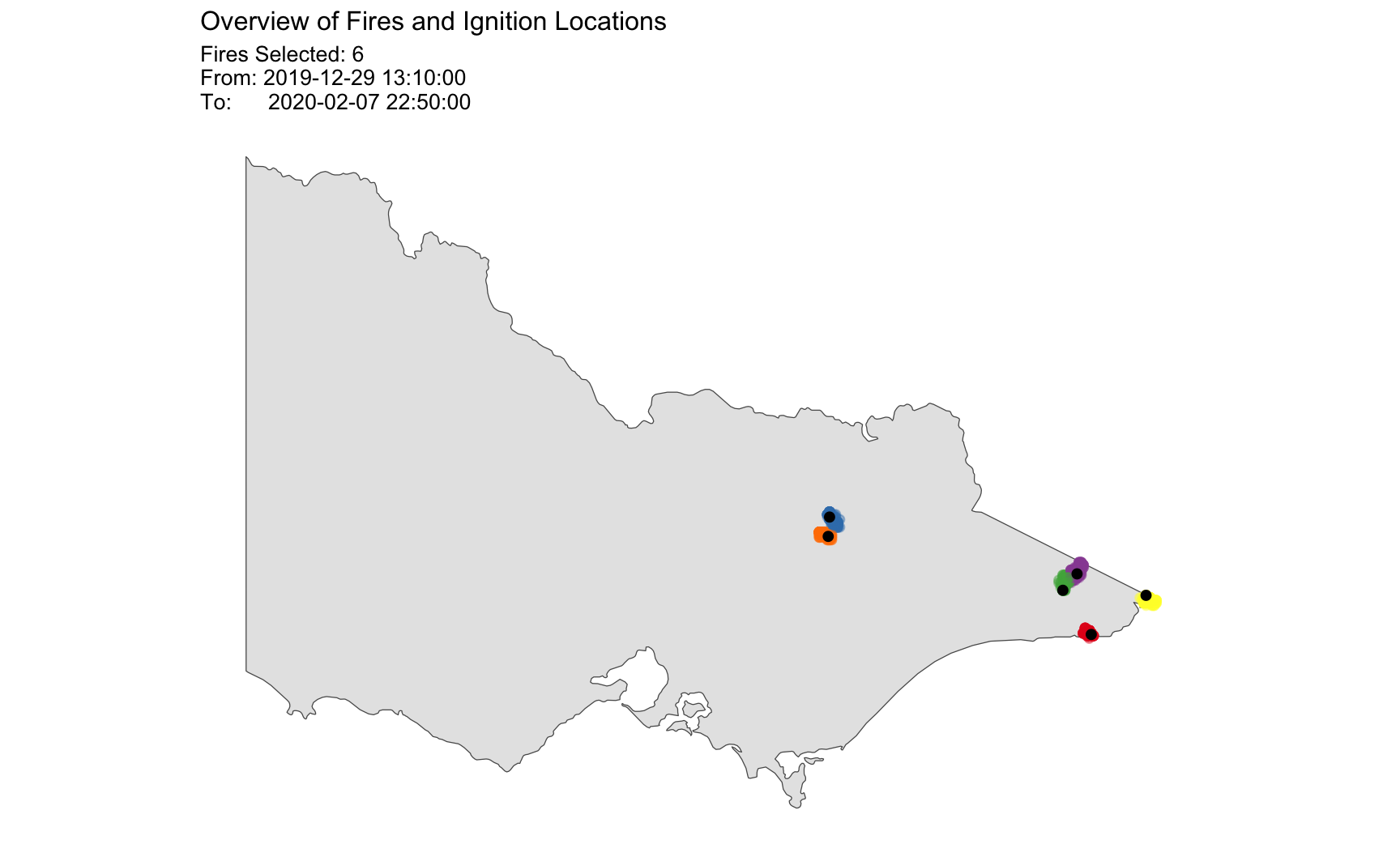} 

}

\caption{This is the default plot for visualizing the spatial distribution of clusters. In the results shown there are six clusters, which correspond to six fires, shown using different colors. The black dots indicate the ignition site for each fire.}\label{fig:demodefplot}
\end{figure}

The default plot allows for visualization of the spatial distribution of the fires. As shown in Figure \ref{fig:demodefplot}, it presents the hotspots of each cluster identified by the algorithm with a single color. The black dot on top of each colored set of points is the ignition location of that fire.
This graphical representation only provides a static view of the hotspot data. To get a dynamic view, users can draw subplots grouped by time.

\begin{verbatim}
plot(result, bg = plot_vic_map())
\end{verbatim}

\hypertarget{fire-movement-plot-for-visualizing-the-fire-dynamics}{%
\paragraph{Fire movement plot for visualizing the fire dynamics}\label{fire-movement-plot-for-visualizing-the-fire-dynamics}}

\begin{figure}

{\centering \includegraphics[width=1\linewidth]{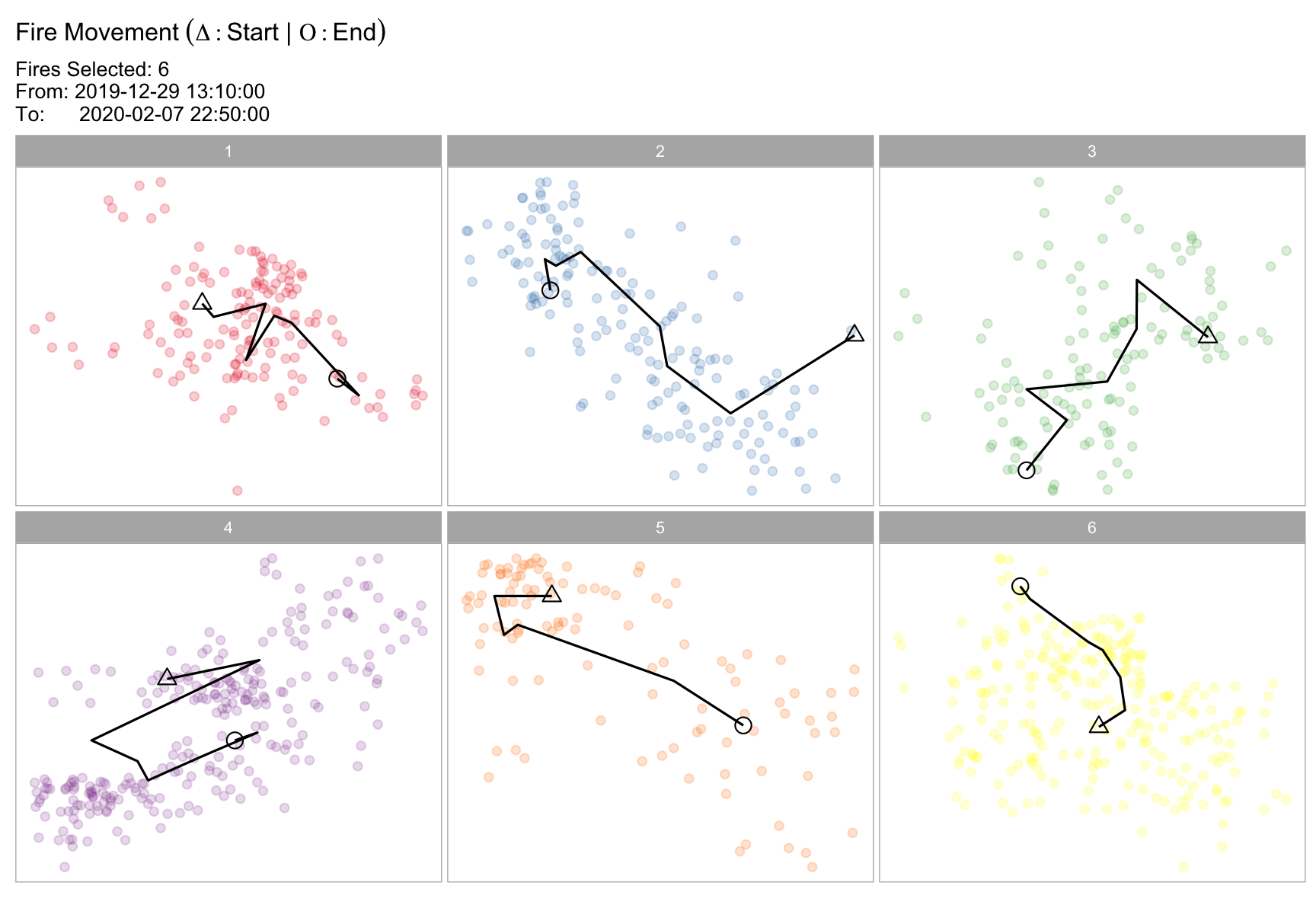} 

}

\caption{This is the fire movement plot for visualizing the fire dynamics. Here there are six clusters, corresponding to six different fires. The path between the ignition point and the end point is drawn with black line, where the triangle is the ignition point and the circle is the end point. (Note that the aspect ratio of the plot reflects the relative spatial ratio of latitude and longitude.)}\label{fig:demomovplot}
\end{figure}

Figure \ref{fig:demomovplot} shows the path of each fire's movement, which is produced by setting the plot argument \texttt{type\ =\ \textquotesingle{}mov\textquotesingle{}}. The fire movement is computed by the \texttt{get\_fire\_mov()} function. Its argument \texttt{step} controls the number of time intervals combined for calculation of the hotspots' centroid for the purpose of this visualization. Using a small value of \texttt{step} will produce a complex path; for example, \texttt{step\ =\ 1} means that the centroid of hotspots in each hour will be calculated to trace the fire's path, while \texttt{step\ =\ 12} combines intervals over the course of half a day.

Note that centroids won't adequately summarize unusual cluster shapes, and it may be necessary to change the visualization to show a full shape with perhaps a convex hull. This would make for a complicated visualization of temporal movement, though. Also if a fire spreads simultaneously in different directions, the centroids would remain in approximately the same locations.

\begin{verbatim}
plot(result, type = "mov", step = 12)
\end{verbatim}

In this plot, hotspots are slightly jittered from their original recorded locations on the spatial grid so repeat observations at a specific latitude and longitude are visible. The triangle identifies the fire's ignition point and the circle is the fire's final location. This provides us with a view into the distribution of hotspots over the course of the fire, as well as the bushfire's overall movement. For example, fire 1 ended southeast of its original ignition location.

\hypertarget{timeline-plot-for-providing-an-overview-of-the-bushfire-season}{%
\paragraph{Timeline plot for providing an overview of the bushfire season}\label{timeline-plot-for-providing-an-overview-of-the-bushfire-season}}

\begin{figure}

{\centering \includegraphics[width=1\linewidth]{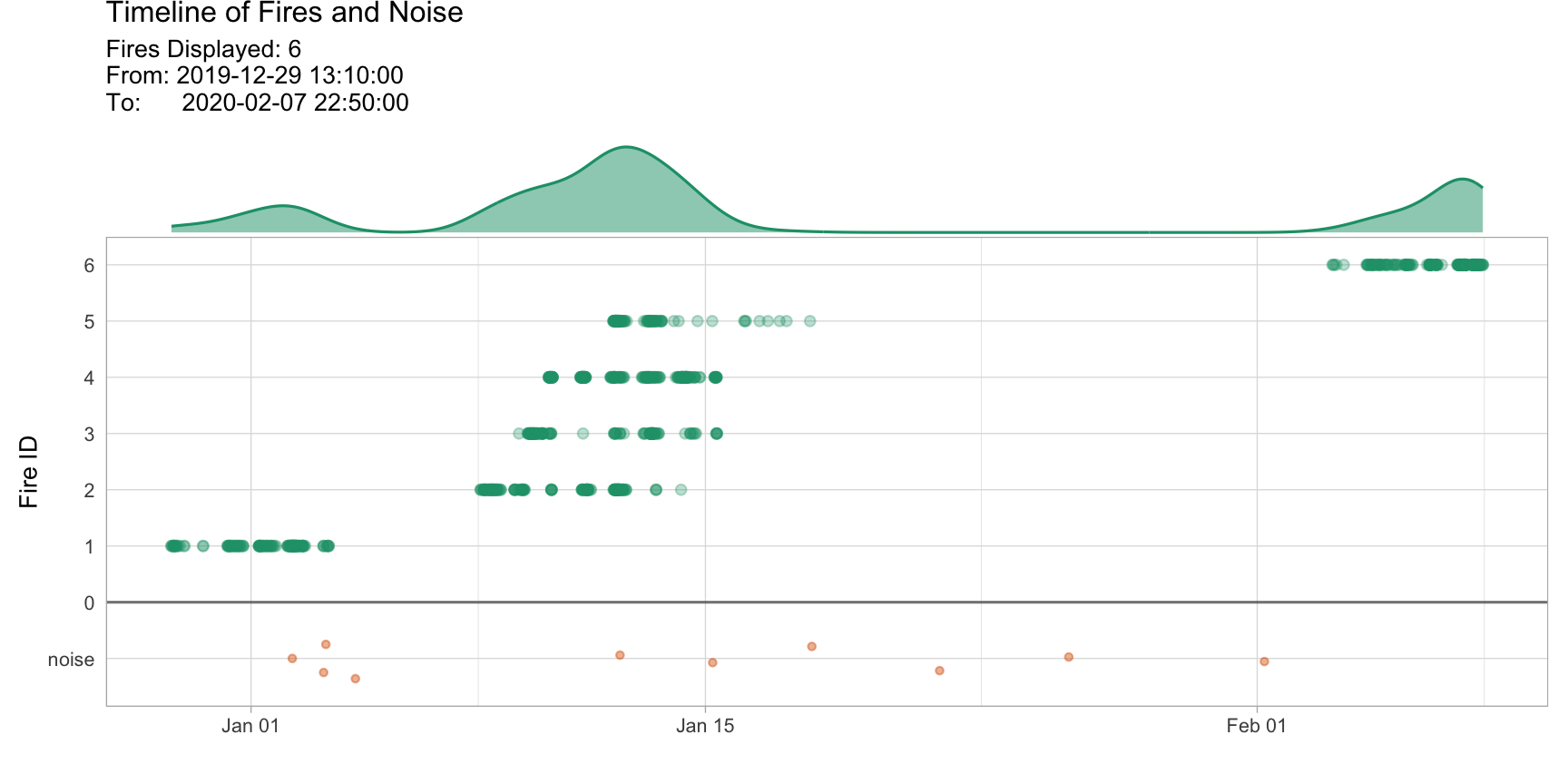} 

}

\caption{This is the timeline plot for providing an overview of the bushfire season. The x-axis is the date and the y-axis is the cluster membership. The observed time of hotspots are shown as dot plots (green). The density plot at the top display the temporal frequency of fire occurrence over the timeframe. The dot plot at the bottom (orange) shows the observed time of hotspots that are considered to be noise.}\label{fig:demotimeline}
\end{figure}

Figure \ref{fig:demotimeline} shows a timeline of fires produced by setting plot argument \texttt{type\ =\ \textquotesingle{}timeline\textquotesingle{}}, which enables study of the intensity of the bushfire season. Hotspots' observed times are plotted along the horizontal line corresponding to their assigned clusters. The green curve along the top captures the temporal density of all fires throughout the season. Noise points are represented with orange dots. The plots demonstrates that the majority of fires in our sample data set burned in mid-January.

\begin{verbatim}
plot(result, type = "timeline")
\end{verbatim}

\hypertarget{application}{%
\subsection{Application}\label{application}}

\begin{figure}

{\centering \includegraphics[width=1\linewidth]{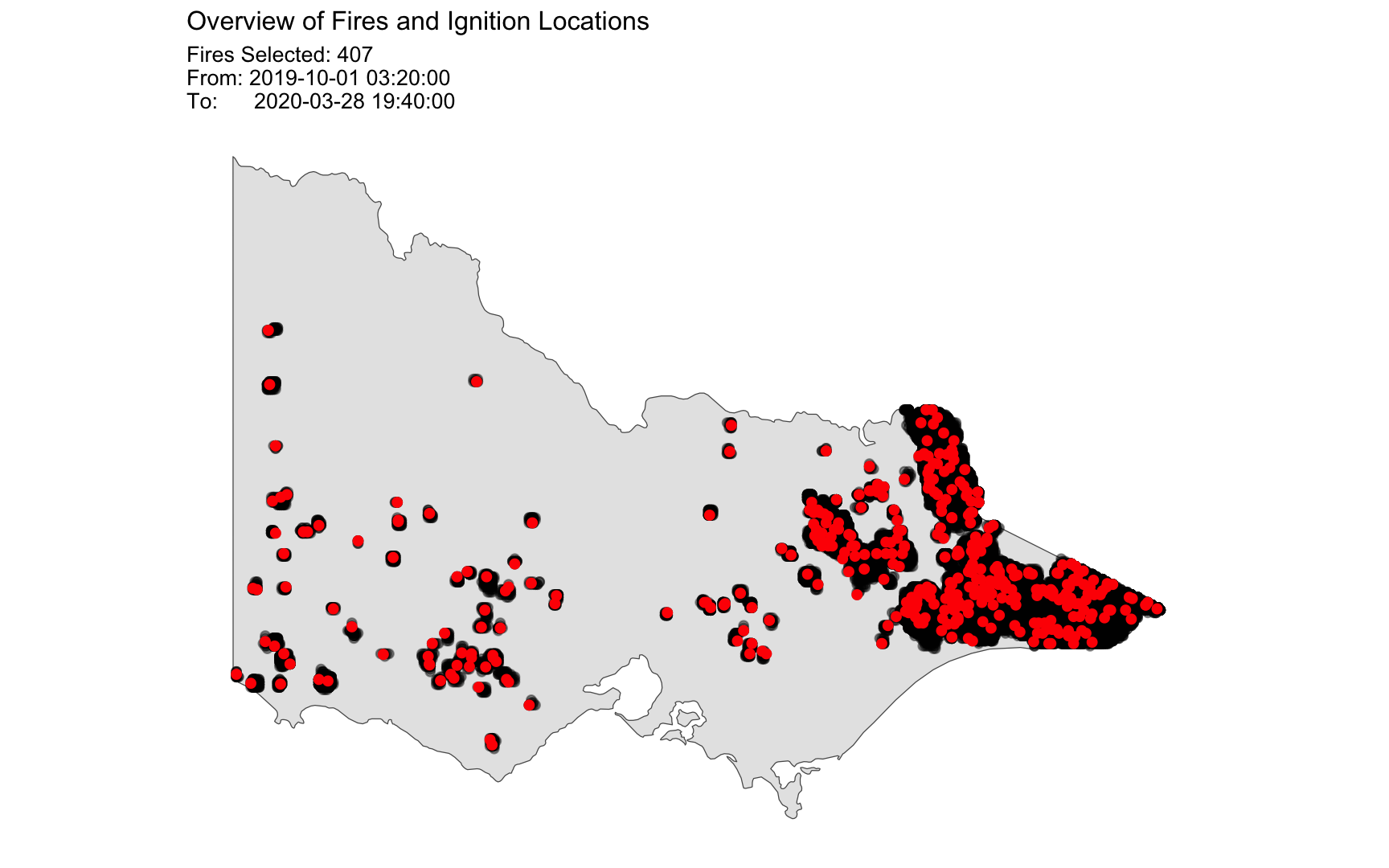} 

}

\caption{ The distribution of hotspots (black) and bushfire ignitions (red) in Victoria during 2019-2020 Australian bushfire season. The spatial distribution of the ignition locations suggest that most of the fires were observed in the east of Victoria.}\label{fig:clusteringfinalresults}
\end{figure}

We illustrate the use of the algorithm by applying it to the entirety of the 75,936 hotspots in Victoria. The spatial distribution of these hotspots for October 2019-March 2020 is shown in Figure \ref{fig:clusteringfinalresults}. This example takes around 5 minutes to run on a 2020 Apple MacBook Pro.

\hypertarget{clustering-the-victorian-hotspot-data}{%
\subsubsection{Clustering the Victorian hotspot data}\label{clustering-the-victorian-hotspot-data}}

To perform the clustering algorithm on the Victorian hotspot data, we first convert the observed time of each hotspot to its corresponding time index by setting the time difference between two successive indices to be \(1\) hour via parameters \texttt{timeStep} and \texttt{timeUnit}. As initially discussed in \protect\hyperlink{algorithm}{Algorithm}, because the original Himawari-8 data is recorded at 10-minute intervals, reassigning each hotspot to its corresponding hourly time index does not cause great loss of information, but serves to significantly shorten the computation time. We then let \texttt{activeTime} be \(24\) time indices and \texttt{adjDist} be \(3000\) meters; see \protect\hyperlink{selecting-parameter-values}{Selecting parameter values} for further discussion of and guidance regarding selection of these choices. Finally, \texttt{minPts} is \(3\) hotspots and \texttt{minTime} is \(3\) time indices.

\begin{verbatim}
result <- hotspot_cluster(hotspots = vic_hotspots,
                          lon = "lon",
                          lat = "lat",
                          obsTime = "obsTime",
                          activeTime = 24,
                          adjDist = 3000,
                          minPts = 4,
                          minTime = 3,
                          timeUnit = "h",
                          timeStep = 1)
\end{verbatim}

The clustering result shows that 407 bushfires are identified among the 75,936 hotspots. This result is reasonable, as historically the number of bushfires reported in Victoria in a year ranges from about \(200\) to \(700\) (Department of Environment, Land, Water \& Planning 2019).

\begin{verbatim}
result
\end{verbatim}

\begin{verbatim}
#> i spotoroo object: 407 clusters | 75936 hot spots (including noise points)
\end{verbatim}

\hypertarget{determining-the-ignition-point-and-time-for-individual-fires}{%
\subsubsection{Determining the ignition point and time for individual fires}\label{determining-the-ignition-point-and-time-for-individual-fires}}

The clustering result provides the ignition location of the bushfire represented by each cluster, and the \texttt{plot()} function now enables us to examine the spatial distribution of bushfire ignitions over the course of the entire summer. The result is given in Figure \ref{fig:clusteringfinalresults}. From this plot, we observe that the large majority of fires burned in eastern Victoria, with a smattering in the south west. These areas are primarily forests and mountains. Very few fires started in or around city of Melbourne, which is located along the southern coast in the middle of the state.

Note that because there are now too many fires to color each of them uniquely (as done previously in Figure \ref{fig:demodefplot}, the default \texttt{plot} function colors each cluster of fires black with their respective ignition points in red. This change in plot output occurs once the number of clusters is greater than nine.

\begin{verbatim}
plot(result, bg = plot_vic_map(), hotspot = TRUE)
\end{verbatim}

\begin{figure}

{\centering \includegraphics[width=1\linewidth]{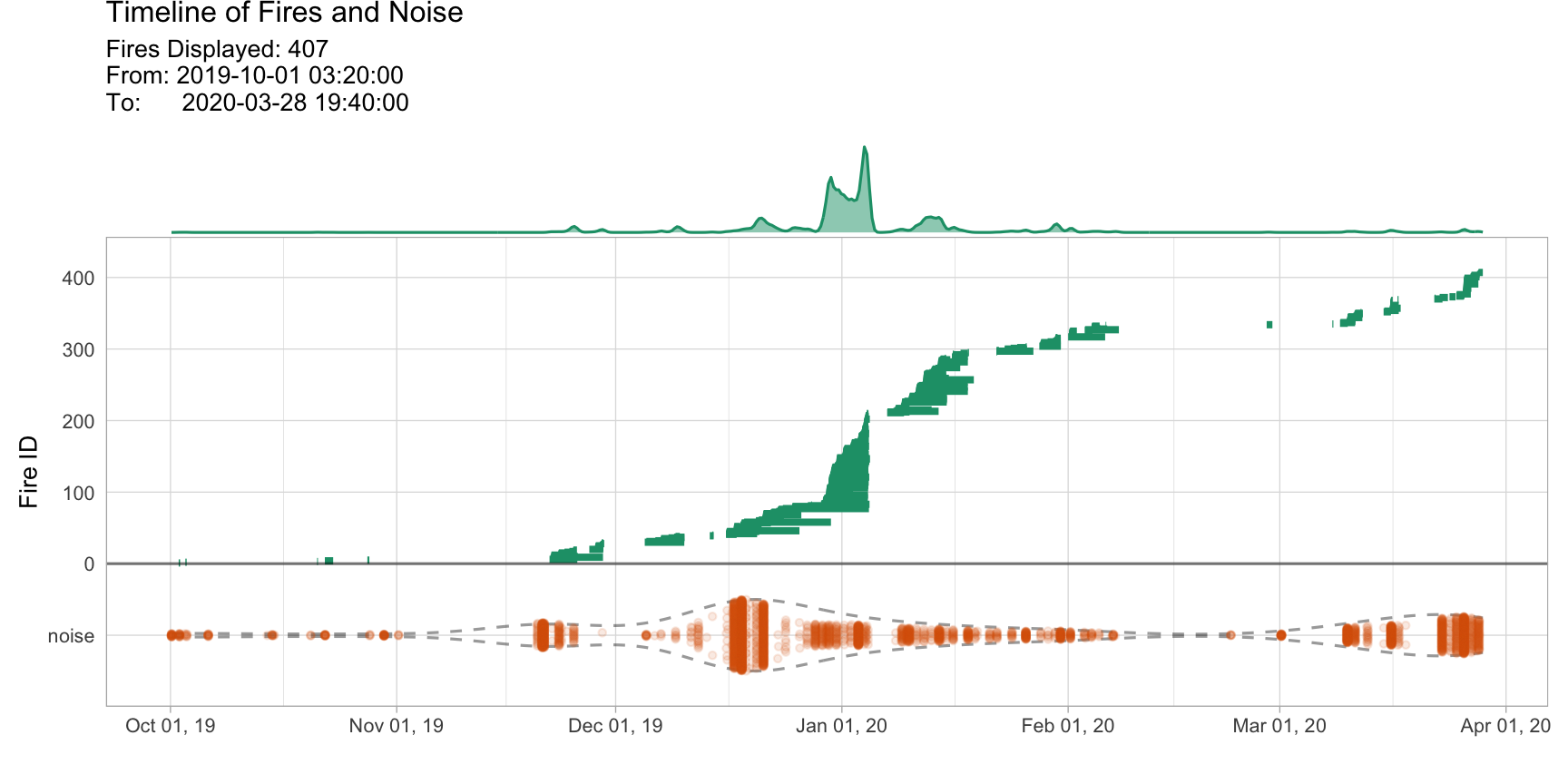} 

}

\caption{Timeline of fires observed in Victoria during the 2019-2020 Australian bushfire season. Clustered hotspots are shown as dotplots (green). The density display of the timeline shows that most fires started in late December and early January. Noise is shown at the bottom (orange), with the dashed lines indicating the density. This plot shows there is a significant number of hotspots that could be considered to be noise, especially in mid-December. It might also suggest that there are lots of short-lived and spatially constrained fires.}\label{fig:himtimeline}
\end{figure}

The ignition time of individual fires and representation of how long they burned is produced with the following code, the output of which is Figure \ref{fig:himtimeline}. This plot shows that the majority of fires were ignited from late December 2019 to early January 2020. We observe a significant number of hotspots identified to be noise points in mid-December, which may imply there are some undetected fire events that were short-lived.

\begin{verbatim}
plot(result, type = "timeline", mainBreak = "1 month", dateLabel = "%b %d, %y")`. 
\end{verbatim}

\hypertarget{tracking-fire-movement}{%
\subsubsection{Tracking fire movement}\label{tracking-fire-movement}}

\begin{figure}

{\centering \includegraphics[width=1\linewidth]{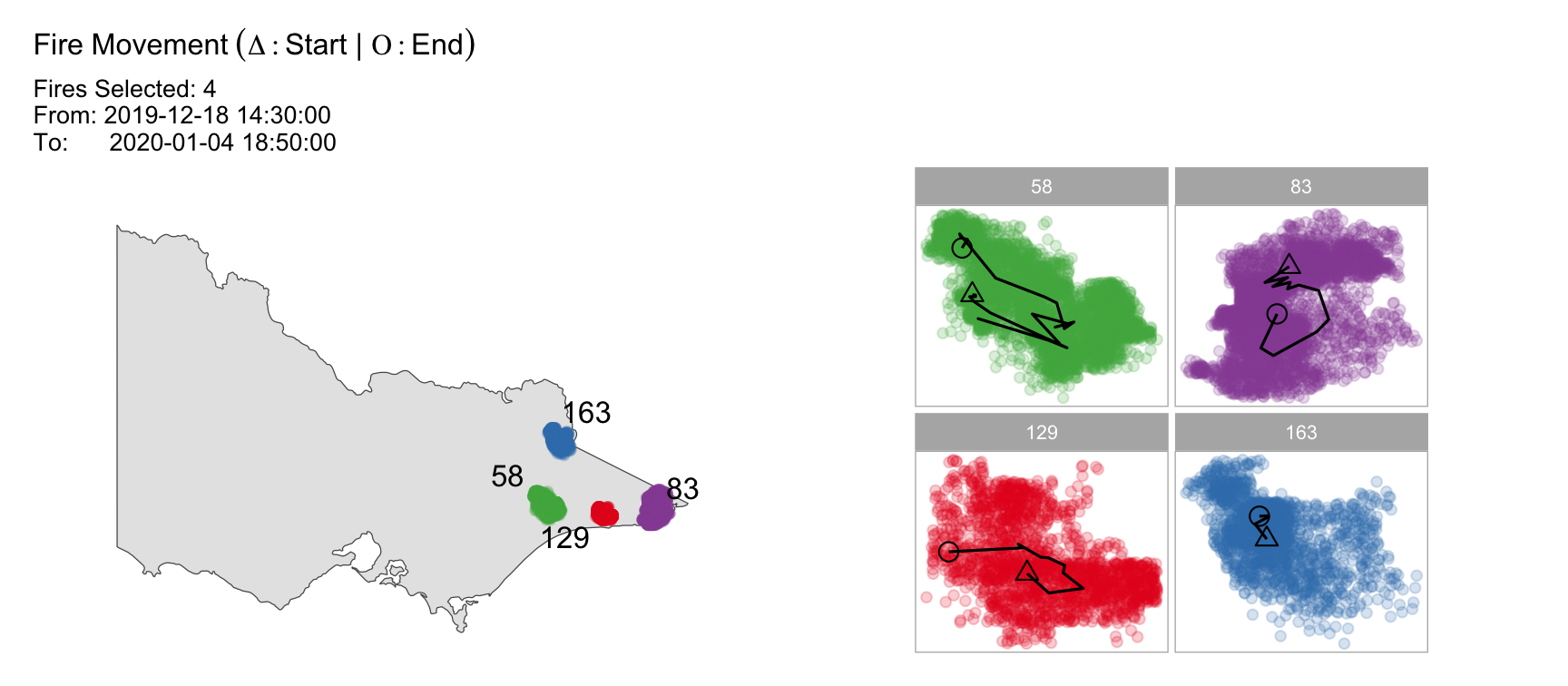} 

}

\caption{Examining the dynamics of the four most intensive fires in Victoria during the 2019-2020 Australian bushfire season. All of the fires covered similar spatial areas over their lifetimes, but the trajectory was quite different. Fire 163 may have spread in many directions simultaneously over the time period, as indicated by the near constant location of the centroid.}\label{fig:firemovem}
\end{figure}

We further study the movement of the four most intensive fires, as identified by the number of hotspots clustered together. These fires burned in the eastern part of Victoria, and the output of the following code is shown in Figure \ref{fig:firemovem}. These fires burned during the time period from December 18th through January 4th. According to the plot, 3 out of the 4 fires moved over the course of their burning, while the center of fire 163 does not appear to move much. This is because simultaneous spread in different directions kept the centroid of the cluster in a similar location over time.

\begin{verbatim}
plot(result, 
     type = "mov", 
     cluster = order(result$ignition$obsInCluster,
                     decreasing = TRUE)[1:4], 
     step = 12, 
     bg = plot_vic_map())
\end{verbatim}

\hypertarget{selecting-parameter-values}{%
\subsubsection{Selecting parameter values}\label{selecting-parameter-values}}

We previously introduced two key parameters, \(adjDist\) and \(activeTime\), in our explanation of the algorithm. While their optimal choice is unknown, a visualization tool enables their tuning such that the user can select reasonable values.

\begin{figure}

{\centering \includegraphics[width=1\linewidth]{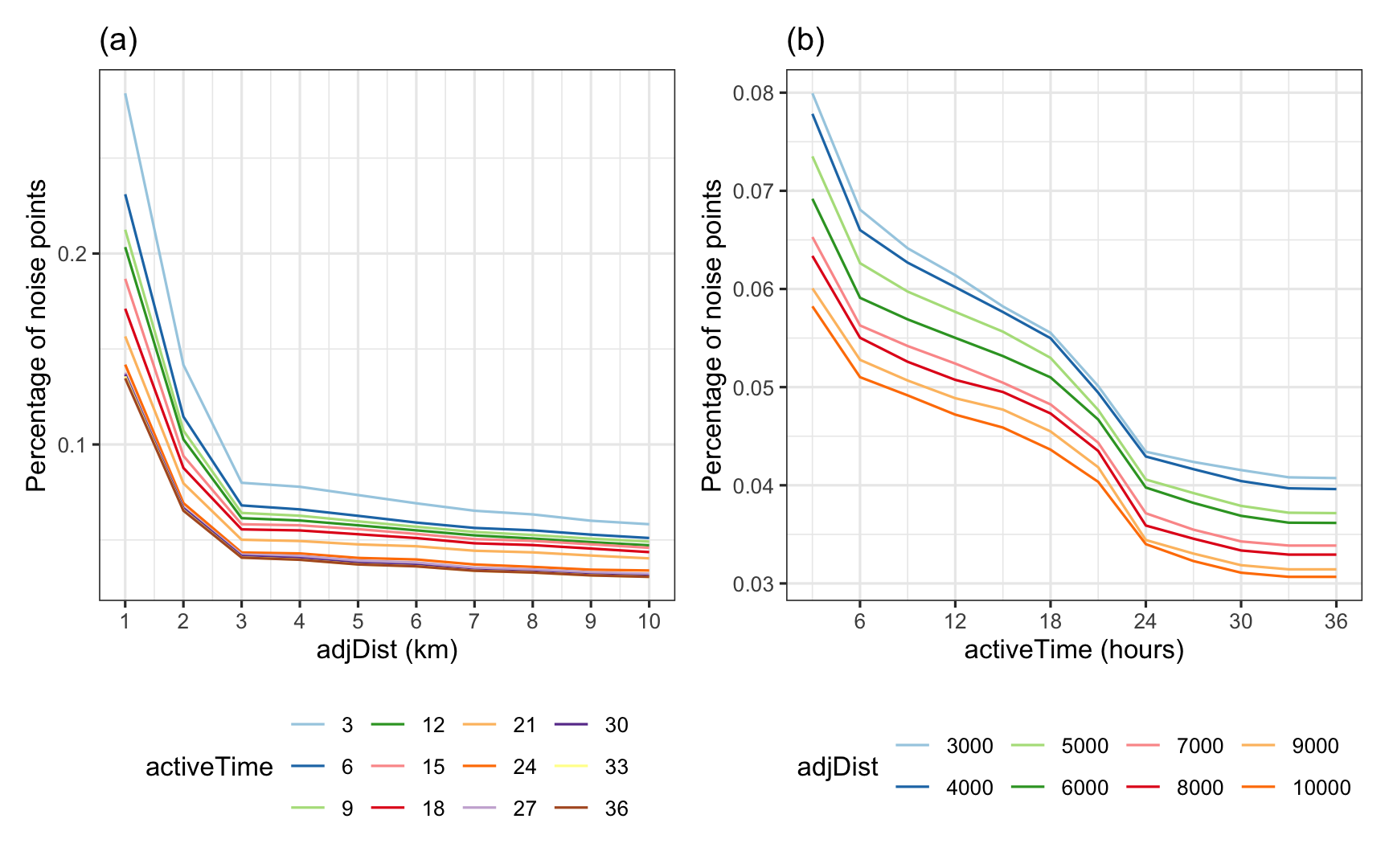} 

}

\caption{Parameter tuning plots, where the best choice of parameter is at a large drop in the percentage of noise points. Here, these are at $AdjDist = 3000$ and $activeTime = 24$. }\label{fig:vis1}
\end{figure}

Consider the relationships between \(adjDist\), \(activeTime\), and the percentage of hotspots observed to be noise points. Increase of either \(adjDist\) or \(activeTime\) represents greater spatial and temporal tolerance, respectively, either of which usually reduces the percentage of noise points. However, if the noise points are spatially and temporally far from the meaningful clusters, increasing either of these two parameters may not significantly reduce the number of clusters. Therefore, to identify what are likely real noise points, we find the smallest values for each of \(adjDist\) and \(activeTime\) after which the percentage of noise points no longer decreases significantly. Based on this logic, we develop a visualization tool inspired by the scree plot (Cattell 1966) used in the principal component analysis and the sorted k-dist plot used in DBSCAN (Ester et al. 1996). Similar to these two plots, users need to determine the values of \(activeTime\) and \(adjDist\) for which the greatest decrease in the percentage of noise points is captured. Figure \ref{fig:vis1} shows the parameter tuning process using this visualization tool. To produce these plots, a grid of parameter values must be evaluated, which can be computationally intensive. For this reason, we do not include this tool in the \CRANpkg{spotoroo}, but leave it for the user to build. The final choice of \(activeTime\) is \(24\) hours and \(adjDist\) is \(3000\) meters.

\hypertarget{conclusions}{%
\subsection{Conclusions}\label{conclusions}}

In this paper, we have proposed a spatiotemporal clustering algorithm to organize satellite hotspot data for the purpose of determining points of bushfire ignition and tracking their movement remotely. This algorithm can be used to cluster hotspots into bushfires of arbitrary shape and size, while following some basic characteristics of bushfire dynamics. It is necessary to tune the parameters of the algorithm. This can be done visually, as demonstrated in the previous section's example application, or determined by expert knowledge of fire dynamics for a geography of interest.

We have provided an implementation of the algorithm in the R package \CRANpkg{spotoroo}, which supports three types of visualizations for presenting different aspects of the bushfire data: the spatial distribution of fires, the movement of individual bushfires, and the timeline of fire ignitions and length of burning. In addition, we have illustrated the use of this package in studying the 2019-2020 Australian bushfire season. The main benefit of this work is that information otherwise infeasible to collect about fires in remote areas can be extracted from satellite hotspot data and processed with the use of this package.

We have previously discussed the ways in which existing algorithms for spatial clustering that incorporate a temporal component are not suitable for our intended purpose. Furthermore, there are no R packages on CRAN that provide implementations of ST-DBSCAN, FSR or MC1 in order to run a direct comparison of computational efficiency or clustering performance. There is an implementation of DBSCAN in R package \CRANpkg{dbscan} that runs extremely fast (\textless{} 10 seconds) on hotspot data from our example application with the proper choice of parameters. However, it clusters over 30\% of the hotspots into one fire, which is not very helpful for tracking fire ignition and spread.

There are possible modifications that could be made to our algorithm for the purpose of improving quality of the clustering result or reduced computation time. We might consider alternative clustering techniques -- for example, grid-based clustering -- at each time window. It may also be possible to use an adaptable \(adjDist\) at different time windows. We leave these extensions for future research.

More complex analyses can be conducted when clustering results are merged with data such as recreation sites, fire stations, roads, weather and vegetation types. An analysis of this type to predict the causes of bushfires in Victoria during the 2019-2020 bushfire season is described in Cook (2020), which yields meaningful suggestions for future fire prevention. We hope others can build upon this work to make use of satellite hotspot data for bushfire research, especially in the context of future bushfire planning and prevention.

\hypertarget{acknowledgements}{%
\subsection{Acknowledgements}\label{acknowledgements}}

This work was made possible through funding by the Australian Centre of Excellence for Mathematical \& Statistical Frontiers, which supported Emily's travel.

This paper was written in \CRANpkg{rmarkdown} with \CRANpkg{knitr} and \CRANpkg{rjtools}. All analyses were performed using R version 4.2.2 (R Core Team 2022). Data were processed with \CRANpkg{readr} and \CRANpkg{dplyr}. Plots were produced with \CRANpkg{ggplot2}, \CRANpkg{ggforce} and \CRANpkg{patchwork}. The code and data to reproduce this paper are available on \href{https://github.com/TengMCing/Hotspots-Clustering-Algorithm}{GitHub}.

\hypertarget{references}{%
\section*{References}\label{references}}
\addcontentsline{toc}{section}{References}

\hypertarget{refs}{}
\begin{CSLReferences}{1}{0}
\leavevmode\vadjust pre{\hypertarget{ref-Abram2021}{}}%
Abram, Nerilie J., Benjamin J. Henley, Alex Sen Gupta, Tanya J. R. Lippmann, Hamish Clarke, Andrew J. Dowdy, Jason J. Sharples, et al. 2021. {``{Connections of Climate Change and Variability to Large and Extreme Forest Fires in Southeast Australia}.''} \emph{{Communications Earth \& Environment}} 2 (1): 8. \url{https://doi.org/10.1038/s43247-020-00065-8}.

\leavevmode\vadjust pre{\hypertarget{ref-climate2021}{}}%
Australian Government Bureau of Meteorology. 2021. {``{Annual Climate Statement 2020}.''} {Australian Government Bureau of Meteorology}. \url{http://www.bom.gov.au/climate/current/annual/aus/}.

\leavevmode\vadjust pre{\hypertarget{ref-stdbscan}{}}%
Birant, Derya, and Alp Kut. 2007. {``{ST-DBSCAN: An Algorithm for Clustering Spatial-Temporal Data}.''} \emph{{Data \& Knowledge Engineering}} 60 (1): 208--21. \url{https://doi.org/10.1016/j.datak.2006.01.013}.

\leavevmode\vadjust pre{\hypertarget{ref-cattell1966scree}{}}%
Cattell, Raymond B. 1966. {``{The Scree Test for the Number of Factors}.''} \emph{{Multivariate Behavioral Research}} 1 (2): 245--76.

\leavevmode\vadjust pre{\hypertarget{ref-conart}{}}%
Cook, Dianne. 2020. {``{Open Data Shows Lightning, Not Arson, Was the Likely Cause of Most Victorian Bushfires Last Summer}.''} \url{https://theconversation.com/open-data-shows-lightning-not-arson-was-the-likely-cause-of-most-victorian-bushfires-last-summer-151912}.

\leavevmode\vadjust pre{\hypertarget{ref-climate2020}{}}%
CSIRO and Australian Government Bureau of Meteorology. 2020. {``{State of the Climate 2020}.''} {CSIRO and Australian Government Bureau of Meteorology}. \url{http://www.bom.gov.au/state-of-the-climate/documents/State-of-the-Climate-2020.pdf}.

\leavevmode\vadjust pre{\hypertarget{ref-Deb2020}{}}%
Deb, Proloy, Hamid Moradkhani, Peyman Abbaszadeh, Anthony S. Kiem, Johanna Engstrom, David Keellings, and Ashish Sharma. 2020. {``{Causes of the Widespread 2019-2020 Australian Bushfire Season}.''} \emph{{Earth's Future}} 8 (11): e2020EF001671. \url{https://doi.org/10.1029/2020EF001671}.

\leavevmode\vadjust pre{\hypertarget{ref-fireorigin}{}}%
Department of Environment, Land, Water \& Planning. 2019. {``{Fire Origins - Current and Historical}.''} \url{https://discover.data.vic.gov.au/dataset/fire-origins-current-and-historical}.

\leavevmode\vadjust pre{\hypertarget{ref-ester1996density}{}}%
Ester, Martin, Hans-Peter Kriegel, Jörg Sander, and Xiaowei Xu. 1996. {``{A Density-Based Algorithm for Discovering Clusters in Large Spatial Databases with Noise}.''} In \emph{{Proceedings of the Second International Conference on Knowledge Discovery and Data Mining}}, 226--31. KDD'96. Portland, Oregon: {AAAI Press}. \url{https://dl.acm.org/doi/10.5555/3001460.3001507}.

\leavevmode\vadjust pre{\hypertarget{ref-Filkov2020}{}}%
Filkov, Alexander I., Tuan Ngo, Stuart Matthews, Simeon Telfer, and Trent D. Penman. 2020. {``{Impact of Australia's Catastrophic 2019/20 Bushfire Season on Communities and Environment. Retrospective Analysis and Current Trends}.''} \emph{{Journal of Safety Science and Resilience}} 1 (1): 44--56. \url{https://doi.org/10.1016/j.jnlssr.2020.06.009}.

\leavevmode\vadjust pre{\hypertarget{ref-Giglio2016}{}}%
Giglio, Louis, Wilfrid Schroeder, and Christopher O. Justice. 2016. {``{The Collection 6 MODIS Active Fire Detection Algorithm and Fire Products}.''} \emph{{Remote Sensing of Environment}} 178: 31--41. \url{https://doi.org/10.1016/j.rse.2016.02.054}.

\leavevmode\vadjust pre{\hypertarget{ref-datamining2012}{}}%
Han, Jiawei, Micheline Kamber, and Jian Pei. 2012. \emph{{Data Mining: Concepts and Techniques, 3rd ed.}} {Morgan Kaufman}. \url{https://www.sciencedirect.com/book/9780123814791/data-mining-concepts-and-techniques}.

\leavevmode\vadjust pre{\hypertarget{ref-hermawati2016}{}}%
Hermawati, Rachma, and Imas Sukaesih Sitanggang. 2016. {``{Web-Based Clustering Application Using Shiny Framework and DBSCAN Algorithm for Hotspots Data in Peatland in Sumatra}.''} \emph{{Procedia Environmental Sciences}} 33: 317--23. \url{https://doi.org/10.1016/j.proenv.2016.03.082}.

\leavevmode\vadjust pre{\hypertarget{ref-Jang2019}{}}%
Jang, Eunna, Yoojin Kang, Jungho Im, Dong-Won Lee, Jongmin Yoon, and Sang-Kyun Kim. 2019. {``{Detection and Monitoring of Forest Fires Using Himawari-8 Geostationary Satellite Data in South Korea}.''} \emph{{Remote Sensing}} 11 (3). \url{https://doi.org/10.3390/rs11030271}.

\leavevmode\vadjust pre{\hypertarget{ref-kalnis2005discovering}{}}%
Kalnis, Panos, Nikos Mamoulis, and Spiridon Bakiras. 2005. {``{On Discovering Moving Clusters in Spatio-Temporal Data}.''} In \emph{{Advances in Spatial and Temporal Databases: 9th International Symposium, SSTD 2005, Angra dos Reis, Brazil, August 22-24, 2005. Proceedings 9}}, 364--81. Springer. \url{https://link.springer.com/chapter/10.1007/11535331_21}.

\leavevmode\vadjust pre{\hypertarget{ref-kisilevich2009spatio}{}}%
Kisilevich, Slava, Florian Mansmann, Mirco Nanni, and Salvatore Rinzivillo. 2009. {``{Spatio-Temporal Clustering}.''} In \emph{{Data Mining and Knowledge Discovery Handbook}}, 855--74. Springer. \url{https://link.springer.com/chapter/10.1007/978-0-387-09823-4_44}.

\leavevmode\vadjust pre{\hypertarget{ref-Kurihara2020}{}}%
Kurihara, Yukio, Kazuhisa Tanada, Hiroshi Murakami, and Misako Kachi. 2020. {``{Australian Bushfire Captured by AHI/Himawari-8 and SGLI/GCOM-C}.''} In \emph{{Proceedings of the JpGU-AGU Joint Meeting}}. \url{https://www.eorc.jaxa.jp/ptree/documents/Poster_H8Wfire_JpGU2020.pdf}.

\leavevmode\vadjust pre{\hypertarget{ref-Loboda2007}{}}%
Loboda, T. V., and I. A. Csiszar. 2007. {``{Reconstruction of Fire Spread Within Wildland Fire Events in Northern Eurasia From the MODIS Active Fire Product}.''} \emph{{Global and Planetary Change}} 56 (3): 258--73. \url{https://doi.org/10.1016/j.gloplacha.2006.07.015}.

\leavevmode\vadjust pre{\hypertarget{ref-nisa2014}{}}%
Nisa, Karlina Khiyarin, Hari Agung Andrianto, and Rahmah Mardhiyyah. 2014. {``{Hotspot Clustering Using DBSCAN Algorithm and Shiny Web Framework}.''} In \emph{{2014 International Conference on Advanced Computer Science and Information System}}, 129--32. \url{https://doi.org/10.1109/ICACSIS.2014.7065840}.

\leavevmode\vadjust pre{\hypertarget{ref-jaxa}{}}%
P-Tree System. 2020. {``{JAXA Himawari Monitor - User's Guide}.''} \url{https://www.eorc.jaxa.jp/ptree/userguide.html}.

\leavevmode\vadjust pre{\hypertarget{ref-r2022}{}}%
R Core Team. 2022. \emph{R: A Language and Environment for Statistical Computing}. Vienna, Austria: R Foundation for Statistical Computing. \url{https://www.R-project.org/}.

\leavevmode\vadjust pre{\hypertarget{ref-Wickramasinghe2016}{}}%
Wickramasinghe, Chathura H., Simon Jones, Karin Reinke, and Luke Wallace. 2016. {``{Development of a Multi-Spatial Resolution Approach to the Surveillance of Active Fire Lines Using Himawari-8}.''} \emph{{Remote Sensing}} 8 (11). \url{https://doi.org/10.3390/rs8110932}.

\leavevmode\vadjust pre{\hypertarget{ref-hotspots}{}}%
Williamson, Grant. 2020. {``{Example Code to Generate Animation Frames of Himawari-8 Hotspots}.''} \url{https://gist.github.com/ozjimbob/80254988922140fec4c06e3a43d069a6}.

\leavevmode\vadjust pre{\hypertarget{ref-Xu2017}{}}%
Xu, Guang, and Xu Zhong. 2017. {``{Real-Time Wildfire Detection and Tracking in Australia Using Geostationary Satellite: Himawari-8}.''} \emph{{Remote Sensing Letters}} 8 (11): 1052--61. \url{https://doi.org/10.1080/2150704X.2017.1350303}.

\end{CSLReferences}

\bibliography{li-dodwell-cook.bib}

\address{%
Weihao Li\\
Monash University\\%
Econometrics and Business Statistics\\
\textit{ORCiD: \href{https://orcid.org/0000-0003-4959-106X}{0000-0003-4959-106X}}\\%
\href{mailto:weihao.li@monash.edu}{\nolinkurl{weihao.li@monash.edu}}%
}

\address{%
Emily Dodwell\\
\\%
\\
\href{mailto:emdodwell@gmail.com}{\nolinkurl{emdodwell@gmail.com}}%
}

\address{%
Dianne Cook\\
Monash University\\%
Econometrics and Business Statistics\\
\url{http://www.dicook.org}\\%
\textit{ORCiD: \href{https://orcid.org/0000-0002-3813-7155}{0000-0002-3813-7155}}\\%
\href{mailto:dicook@monash.edu}{\nolinkurl{dicook@monash.edu}}%
}